\documentclass[11pt,a4paper]{article}
\usepackage[hyperref]{emnlp2020}
\usepackage{times}
\usepackage{latexsym}
\usepackage{amsfonts}
\usepackage{amsmath,amssymb}
\usepackage{subfigure}
\usepackage{graphicx}

\usepackage{tabularx}
\usepackage{algorithm}
\usepackage{algorithmicx}
\usepackage{algpseudocode}
\usepackage{multirow,makecell}
\usepackage{booktabs}
\usepackage{xcolor}

\usepackage{microtype}
\aclfinalcopy 

\title{Learning to Attack: Towards Textual Adversarial Attacking in Real-world Situations}

\author{Yuan Zang$^{12}$\thanks{\ \ Indicates equal contribution.}\hspace{0.3em}, 
Bairu Hou$^{23*}$\hspace{0.2em}, 
Fanchao Qi$^{12*}$, 
Zhiyuan Liu$^{12}$\hspace{0.2em}, \\
{\bf Xiaojun Meng$^{4}$,
Maosong Sun$^{12}$}\\
$^{1}$Department of Computer Science and Technology, Tsinghua University \\
$^{2}$Institute for Artificial Intelligence, Tsinghua University \\
Beijing National Research Center for
Information Science and Technology\\
$^{3}$School of Economics and Management, Tsinghua University \quad
$^{4}$Huawei Noah's Ark Lab\\
{\tt \{zangy17,qfc17\}@mails.tsinghua.edu.cn, houbr.17@sem.tsinghua.edu.cn}\\
{\tt \{liuzy,sms\}@tsinghua.edu.cn, mengxiaojun2@huawei.com}
}

\date{}

\begin{document}
\maketitle
\begin{abstract}
Adversarial attacking aims to fool deep neural networks with adversarial examples.
In the field of natural language processing, various textual adversarial attack models have been proposed, varying in the accessibility to the victim model.
Among them, the attack models that only require the output of the victim model are more fit for real-world situations of adversarial attacking.
However, to achieve high attack performance, these models usually need to query the victim model too many times, which is neither efficient nor viable in practice.
To tackle this problem, we propose a reinforcement learning based attack model, which can learn from attack history and launch attacks more efficiently.
In experiments, we evaluate our model by attacking several state-of-the-art models on the benchmark datasets of multiple tasks including sentiment analysis, text classification and natural language inference.
Experimental results demonstrate that our model consistently achieves both better attack performance and higher efficiency than recently proposed baseline methods.
We also find our attack model can bring more robustness improvement to the victim model by adversarial training.
All the code and data of this paper will be made public.

\end{abstract}

\section{Introduction}

Deep neural networks (DNNs) have been proved vulnerable to adversarial attacks, which maliciously craft \textit{adversarial examples} to fool the victim model \citep{szegedy2014intriguing,goodfellow2015explaining}.
For instance, highly poisonous phrases with minor modification can easily deceive Google's toxic comment detection system \citep{hosseini2017deceiving}.
With the broad use of DNN-based natural language processing (NLP) systems, such as spam filtering \citep{bhowmick2018mail} and malware detection \citep{mclaughlin2017deep}, there is growing concern about their security.
As a result, research into textual adversarial attacking becomes increasingly important.

In recent years plenty of adversarial attack models have been proposed \citep{zhang2019generating}. 
Nevertheless, few of them work satisfactorily in real-world attack situations.
Existing adversarial attack models can be roughly classified into four categories according to the accessibility to the victim model: gradient-based, score-based, decision-based and blind models.
First, gradient-based models, also known as white-box models, require full knowledge of the victim model to perform gradient computation \citep{papernot2016crafting,ebrahimi2018hotflip}.
Unfortunately, we hardly know the architecture of the victim model in real-world attack situations, let alone compute the gradients.

Second, blind models do not need to know anything about the victim model, but their attack performance is usually not good enough, precisely because of complete ignorance about the victim model. 
Specifically, existing blind models either implement character-level random perturbations \citep{ebrahimi2018hotflip} or conduct sentence-level distracting \citep{jia2017adversarial} and paraphrasing \citep{iyyer2018adversarial}.
However, character-level attacks are easy to repulse \citep{pruthi2019combating}, and sentence-level attacks cannot guarantee attack validity, i.e, keeping the ground-truth label of the adversarial example the same as original input.
More importantly, the attack success rates of most blind models are unsatisfactory.

Finally, score- and decision-based attack models seem to be more suitable for real-world adversarial attack situations.
They only need to know the output of the victim models -- the former requires prediction scores and the latter just needs the final prediction decision. 
Existing score- and decision-based attack models have achieved great attack performance \citep{alzantot2018generating,zhao2018generating,jin2019bert,zang2020word}, but they have a significant problem.
To craft an adversarial example, these models have to iteratively make perturbations and query the victim model too many times, e.g., a very recent score-based model needs to query the victim model more than $2,000$ times on average to generate an adversarial example \citep{zang2020word}.
It is neither efficient nor practical to invoke the victim model so many times in real-world situations of adversarial attacking.

We argue that the low efficiency of existing score- and decision-based attack models results from that they have no learning ability and simply follow certain fixed optimization rules to attack, e.g., greedy algorithm \citep{jin2019bert}, genetic algorithm \citep{alzantot2018generating} and particle swarm optimization \citep{zang2020word}. 

To solve this problem, we propose to build an attack model possessing learning ability, which can learn lessons from attack history and store them in its parameters so as to improve attack efficiency.
Considering no labeled data are available in adversarial attacking, we design our model following the reinforcement learning paradigm.
There are two main operations in our model, including identifying key words in the original sentences that crucially influence the decision of the victim model, and selecting appropriate substitutes to replace them.
Our model is aimed at learning an optimal policy under which a series of substitution operations are iteratively conducted to generate adversarial examples. 

In experiments, we evaluate our attack model on the benchmark datasets of three typical NLP tasks including sentiment analysis, text classification and natural language inference.
The victim models are respective (nearly) state-of-the-art models of the datasets, namely ALBERT \citep{lan2019albert}, XLNet \citep{yang2019xlnet} and RoBERTa \citep{liu2019roberta}, and two open APIs.
Since our model can work in both score- and decision-based attack settings, we carry out experiments in the two settings.
Experimental results show that our attack model consistently outperforms the baseline methods on all the datasets in terms of both attack success rate and attack efficiency.
We also find our model can bring more robustness improvement to the victim model by adversarial training.


\section{Related Work}
As mentioned above, textual adversarial attack models can be categorized into four classes according to the accessibility to victim model, namely gradient-based, score-based, decision-based and blind models.

Gradient-based attack models, also named white-box attack models, require full knowledge of the victim model to calculate the gradient with respect to the model input.
They are mostly inspired by the fast gradient sign method \citep{goodfellow2015explaining} and forward derivative method \citep{papernot2016crafting} of adversarial attacking on computer vision, and make some adaptations to the discrete text \citep{papernot2016crafting,sato2018interpretable,liang2018deep,ebrahimi2018hotflip,wallace2019universal}.
In real-world attack situations, gradient-based models are hardly able to work because the victim model is usually not fully accessible. 

In contrast to white-box attack models, black-box attack models do not need to possess complete knowledge of the victim model, and can be subclassified into score-based, decision-based and blind models.
Blind models are ignorant of the victim model at all.
Therefore, they can only make arbitrary perturbations, e.g., adding relevant sentences \citep{jia2017adversarial}, paraphrasing \citep{iyyer2018adversarial} and random character-level perturbations including substitution, deletion and swapping \citep{belinkov2018synthetic,gao2018black}.
These models are hard to achieve high attack performance.
And they tend to change the ground-truth label of the original input or can be easily resisted \citep{pruthi2019combating}.

Score- and decision-based attack models are more fit for realistic adversarial attack situations.
Score-based models rely on the prediction scores of the victim model, e.g., classification probabilities of all the classes, while decision-based models require the final decision of the victim model only, e.g., the predicted class.
Quite a lot of attack models are score-based, e.g., word substitution attack models based on genetic algorithm \citep{alzantot2018generating}, greedy algorithm \citep{ren2019generating,jin2019bert}, Metropolis-Hastings sampling \citep{zhang2019generating} and particle swarm optimization \citep{zang2020word}, visually similar character substitution model \citep{eger2019text}, and human-in-the-loop adversarial writing model \citep{wallace2019trick}.
Only a few of existing attack models are decision-based, e.g., performing perturbations in the continuous latent semantic space \citep{zhao2018generating} and rewriting sentences following semantically equivalent adversarial rules \citep{ribeiro2018semantically}.
Most of score- and decision-based models utilize the output of the victim model (prediction scores or final decision) as guiding signals and adjust the perturbations iteratively.
Although effective, they suffer serious problem of efficiency. 
To obtain a high attack success rate, these models normally need to query the victim model thousands of times, which is actually impractical in real-world adversarial attack situations.

\section{Methodology}
In this section, we delineate our attack model, which can work in both score- and decision-based attack settings. 
We first present an overview of our model in §\ref{sec:method1}, and then we detail our model in the score- and decision-based attack settings in §\ref{sec:method2} and §\ref{sec:method3} respectively.

\subsection{Model Overview}
\label{sec:method1}
Our attack model is designed for word-level attacks, which have been considered to have better overall attack performance as compared to character- and sentence-level attacks \citep{wang2019natural,zang2020word}.
There are two main operations in our attack model.


(1) The first one is to identify key words to be substituted in the original sentence.
Previous work has demonstrated that prediction results of neural network-based text classification models are highly dependent on several key words \citep{li2016understanding}.
Therefore, it would be effectual to find and substitute the key words in the original input in adversarial attacks.

(2) The second operation is to select appropriate substitutes to replace the identified key words.
Here a set of candidate substitutes have already been prepared for each word in the original sentence, which can be generated by some candidate substitute nomination methods such as synonym substitution. 
Theoretically, our attack model can be combined with any candidate substitute nomination method.
In experiments, we combine three representative ones with our attack model for evaluation.

By making a substitution, a potential adversarial example is generated, and we can use it to attack the victim model.
If successful, i.e., the victim model yields a prediction result that is different from the ground-truth label of the the original input, 
we terminate this process and output the adversarial example, otherwise we repeat above steps to iteratively make substitutions.

Seeing no labeled data are available in adversarial attacking, to endow our attack model with learning ability, we adopt the reinforcement learning paradigm.
Specifically, we regard the above two operations as the \textit{action}, and our goal is to learn a \textit{policy} under which an adversarial example is generated by taking a series of actions (substitution operations).
Next we give a brief introduction to policy gradient \citep{sutton2000policy}, the method we use for policy learning.


\subsubsection*{Brief Introduction to Policy Gradient}
In reinforcement learning, an agent is expected to learn an optimal policy to earn maximum rewards in a Markov decision process \citep{sutton2018reinforcement}.
Each step of the process $t \in \{0,1,\cdots,T-1\}$ can be described by the triplet of state ${s_t}$, action ${a_t}$ and reward ${r_{t}}$.
Total rewards of the whole process with respect to a parameterized policy $\boldsymbol{\theta}$ is
\begin{equation}
\setlength{\abovedisplayshortskip}{3pt}
\setlength{\abovedisplayskip}{3pt}
\setlength{\belowdisplayshortskip}{3pt}
\setlength{\belowdisplayskip}{3pt}
    {J}(\boldsymbol{\theta})=\mathbb{E}(\sum_{t=0}^{T-1}{\gamma^t{r_{t}}}|\boldsymbol{\theta}),
\end{equation}
where $\gamma\in[0,1]$ is the discount factor weighting future  rewards less than immediate rewards.

Policy gradient optimizes the policy $\boldsymbol{\theta}$ by gradient ascent:
\begin{equation}
\setlength{\abovedisplayshortskip}{2pt}
\setlength{\abovedisplayskip}{2pt}
\setlength{\belowdisplayshortskip}{3pt}
\setlength{\belowdisplayskip}{3pt}
    \boldsymbol{\theta}\leftarrow\boldsymbol{\theta}+\alpha\nabla_{\boldsymbol{\theta}} J(\boldsymbol{\theta}),
\label{eq:1}
\end{equation}
where $\alpha$ is the learning rate.
To calculate $\nabla_{\boldsymbol{\theta}} J(\boldsymbol{\theta})$, a common algorithm is REINFORCE \citep{williams1992simple,sutton2018reinforcement}.
It is based on the Monte Carlo method and approximately calculates $\nabla_{\boldsymbol{\theta}} J(\boldsymbol{\theta})$ by
\begin{equation}
\setlength{\abovedisplayshortskip}{2pt}
\setlength{\abovedisplayskip}{2pt}
\setlength{\belowdisplayshortskip}{3pt}
\setlength{\belowdisplayskip}{3pt}
\begin{aligned}
    \nabla_{\boldsymbol{\theta}}J(\boldsymbol{\theta}) 
    &\sim\sum_{t=0}^{T-1}\nabla_{\boldsymbol{\theta}} \log \pi_{\boldsymbol{\theta}}(a_t|s_t) G_t,
\end{aligned}
\label{eq:2}
\end{equation}
where $\pi_{\boldsymbol{\theta}}({a_t}|{s_t})$ is the probability of taking action $a_t$ in state $s_t$ under the policy $\boldsymbol{\theta}$, and
${G_t}=\sum_{t'=t}^{T-1}{\gamma^{t'-t}r_{t'}}$.

\subsection{Score-based Attacking}
\label{sec:method2}
In this subsection, we describe our attack model in the score-based attack setting, in which the prediction scores of the victim model with respect to any input are accessible.

As mentioned above, our attack model is supposed to learn a policy which directs the two actions including key word identification and substitute selection.
We parameterize the policy with two sets of probability vectors which are related to the two actions respectively.
Next we present our model step by step.


\paragraph{Policy Initialization}
Suppose $x$ is the original sentence that we want to perturb and has $m$ words, 
i.e., $x=w_1w_2\cdots w_m$, we first design an $m$-dimensional key word identification probability vector $\mathbf{p}^{x}$, whose $i$-th dimension is the probability that $w_i$ is identified as a key word and substituted.
Each dimension of $\mathbf{p}^{x}$ is initialized to $\frac{1}{m}$.
For each $w_i$, we design an $n_i$-dimensional substitute selection probability vector $\mathbf{q}^x_i$, where $n_i$ is the number of the candidate substitutes of $w_i$.
The $j$-th dimension of $\mathbf{q}^x_i$ is the probability that $w_i$ is replaced by its $j$-th candidate substitute.
We initialize each dimension of $\mathbf{q}^x_i$ with $\frac{1}{n_i}$.
Here $\mathbf{p}^{x}$ and $\mathbf{Q}^x=\{\mathbf{q}^x_1,\cdots,\mathbf{q}^x_m\}$ constitute the parameters of the policy.

\paragraph{Sampling}
For a given original sentence $x$,
we first implement sampling without replacement from the probability distribution $\mathbf{p}^x$ to obtain
$T=\lfloor {\delta}m \rfloor$ words to be substituted, where ${\delta}$ is the maximum modification rate stipulating the upper limit of the proportion of substituted words.
These words to be substituted form a set $\mathbb{S}$.
Then for each $w_s\in\mathbb{S}$, we sample one word $w_s'$ from the probability distribution of its candidate substitutes $\mathbf{q}^x_s$.
Next we iteratively substitute all the sampled words $w_s\in\mathbb{S}$ with $w_s'$ one by one.
Each substitution can generate a potential adversarial example, and there are $T$ potential adversarial examples in total.
If one of them induces the victim model to yield a prediction result different from the ground-truth label of the original input $x$, the algorithm terminates and outputs that adversarial example.
Otherwise, we move on to calculating the reward of each substitution to update parameters.

\paragraph{Reward Calculation} 
We desire the victim model not to predict the original sentence's ground-truth label as much as possible.
Hence, we define the reward of the $t$-th substitution $r_t$ as the decrements of the prediction score of the ground-truth label.
Specifically,
\begin{equation}
\setlength{\abovedisplayshortskip}{3pt}
\setlength{\abovedisplayskip}{3pt}
\setlength{\belowdisplayshortskip}{3pt}
\setlength{\belowdisplayskip}{3pt}
    {r_t}=P(y_g|x)-P(y_g|{x_t}),
\label{eq_reward}
\end{equation}
where $x_t$ is the generated potential adversarial example after the $t$-th substitution, and $P(y_g|x)$ is the prediction score of the original sentence's ground-truth label given by the victim model.


\paragraph{Policy Updating}
We use policy gradient to update $\boldsymbol{\theta} = (\mathbf{p}^x,\mathbf{Q}^x)$, as in Equation \eqref{eq:1} and \eqref{eq:2}.
Afterwards, the algorithm goes back to the \textbf{Sampling} step to re-generate potential adversarial examples.

\subsection{Decision-based Attacking}
\label{sec:method3}
In the decision-based attack setting, prediction scores of the victim model are unavailable.
Therefore, we cannot calculate the reward and optimize the policy as in the score-based attack setting. 
A simple workaround is to set the reward in Equation \eqref{eq_reward} to a negative constant if the substitution does not generate an adversarial example.
Although workable, the attack performance would be very limited.
To improve attack performance, we also make adaptations of policy initialization.

Previous work has proved the transferability of adversarial examples \citep{goodfellow2015explaining}, which means an adversarial example designed for a victim model is also likely to fool another victim model.
We aim to utilize this transferability by initializing a decision-based attack model with the policy of a \textit{score-based} attack model that has been pre-trained to attack another virtual victim model.
However, original parameterization restricts the transferability of policy, i.e., each original sentence $x$ has a specific policy parameterized by $\mathbf{p}^{x}$ and $\mathbf{Q}^x$ which cannot be transferred.
To solve this problem, we adjust the parameterization of the policy for the pre-trained score-based attack model.


For the probability distribution of key word identification, instead of directly using a probability vector, we design a regression model $f$ to learn it:
\begin{equation}
\setlength{\abovedisplayshortskip}{3pt}
\setlength{\abovedisplayskip}{3pt}
\setlength{\belowdisplayshortskip}{3pt}
\setlength{\belowdisplayskip}{3pt}
    \mathbf{p}^x=f_{\boldsymbol{\vartheta}}(x),
\end{equation}
where $\boldsymbol{\vartheta}$ denotes the parameters of $f$.
We choose BERT \citep{devlin2019bert} plus a multi-layer perceptron as the regression model. 
During attacking we freeze BERT and all the learnable parameters are from the multi-layer perceptron.


For the parameter related to substitute selection, we make the probability vector associated with the target word itself only, 
i.e., 
$\mathbf{Q}^x \rightarrow\mathbf{Q}=\{\mathbf{q}_w|w\in\mathbb{V}\}$, where $\mathbb{V}$ is the vocabulary.

With these adjustments, we train a score-based attack model by attacking a virtual victim model with accessible prediction scores.
Since we have no idea of the architecture of the true victim model, we can simply choose a popular classification model such as LSTM \citep{hochreiter1997long} and BERT as the virtual victim model.
The policy parameters $\boldsymbol{\vartheta}$ and $\mathbf{Q}$ are updated by policy gradient and finally stored.

\begin{table*}[!tb]
    \centering
    \resizebox{.98\linewidth}{!}{
    \begin{tabular}{cccccccc}
        \toprule
        Dataset & Task & \#Classes  & \#Train & \#Validation & \#Test & Victim Model & Accuracy \\
        \midrule
        SST-2 & Sentiment Analysis & 2  & 6,920 & 872 & 1,821 & ALBERT\ /\ MA\ /\ MC & 94.49\ /\ 71.61\ /\ 72.75\\
        AG News & Text Classification&4  & 96,000 & 24,000 & 7,600& XLNet & 94.67\\
        MNLI-m & Natural Language Inference & 3  & 382,885 & 9,817 & 9,815& RoBERTa & 89.28\\
        \bottomrule
    \end{tabular}
    }
    \caption{Details of three evaluation datasets and the accuracy results of corresponding victim models. MA and MC denote Microsoft Azure and Meaning Cloud APIs respectively. }
\label{tab:dataset}
\end{table*}

Afterwards, we use the policy parameters of the trained score-based attack model to initialize the policy of the decision-based attack model.
And then we iteratively conduct sampling, reward calculation and policy update until finding an adversarial example.\footnote{Theoretically we can also use a pre-trained model to initialize the policy of a \textit{score-based} attack model, but we find the performance boost is marginal as compared with the extra cost of time in experiments. Besides, the adjusted policy parameterization is only used in the pre-trained attack model because it needs to update much more parameters ($\boldsymbol{\vartheta}$ vs. $\mathbf{p}^x$) when attacking one instance which is quite inefficient for real adversarial attacks.
}

\section{Experiments}
In this section, we empirically evaluate our attack model on three typical NLP tasks 
including sentiment analysis, text classification, and natural language inference 
in both score- and decision-based attack settings.


\subsection{Datasets and Victim Models}
For each task, we choose one representative benchmark dataset for evaluation and one (nearly) state-of-the-art model as the victim model.
To improve the diversity and extensiveness of the evaluation, we intentionally choose three different victim models.
All of them are popular large pre-trained models and have powerful performance across many language understanding tasks.

(1) For {sentiment analysis}, we use the SST-2 dataset \citep{socher2013recursive}, which is composed of sentences with ``positive'' or ``negative'' labels.
The victim model we choose is \textbf{ALBERT} \citep{lan2019albert}, whose specific version is ALBERT-xxlarge.

(2) For {text classification}, we utilize AG News corpus \citep{zhang2015character} as the evaluation dataset, which comprises news articles that are categorized into four classes: World, Sports, Business, and Sci/Tech. 
We use \textbf{XLNet} \citep{yang2019xlnet}, specifically XLNet-large-cased, as the victim model.

(3) For {natural language inference}, we use the matched MNLI dataset \citep{williams2017broad} (MNLI-m), in which each instance is composed of a sentence pair (premise and hypothesis) and a relational label (entailment, contradiction or neutral).
The victim model on this task is \textbf{RoBERTa} \citep{liu2019roberta} (RoBERTa-large).


To evaluate our model in a more realistic situation, we also choose two open APIs as the victim models on sentiment analysis, namely 
Microsoft Azure Text Analytics\footnote{\url{https://azure.microsoft.com/en-us/services/cognitive-services/text-analytics/}} for score-based attacking, and the Meaning Cloud sentiment analysis API\footnote{\url{https://www.meaningcloud.com/products/sentiment-analysis}} for decision-based attacking. 
The evaluation dataset is still SST-2.
Details of the datasets and performance of the victim models are shown in Table \ref{tab:dataset}.

\subsection{Baseline Methods}
\subsubsection*{Score-based Baseline Methods}
There are many score-based attack models, and we select three representative word-level score-based models as the baselines, which utilize different attack algorithms and candidate substitute nomination methods.


\textbf{GA+Embedding}.
This attack model \citep{alzantot2018generating} adopts the genetic algorithm (GA) to generate adversarial examples and uses word embedding distance to determine candidate substitutes.

\textbf{PWWS+Synonym}. 
This baseline model \citep{ren2019generating} employs the greedy algorithm based on the probability weighted word saliency (PWWS) and nominates synonyms from WordNet \citep{miller1995wordnet} as candidate substitutes.


\textbf{PSO+Sememe}.
This model \citep{zang2020word} is based on the particle swarm optimization (PSO) algorithm \citep{eberhart1995particle}, where the candidate substitutes are nominated via a kind of special {sememe} knowledge \citep{bloomfield1926set}.
Previous experimental results showed that this model achieved state-of-the-art attack results.

To ensure evaluation fairness, we combine our attack model (denoted by \textbf{RL}) with the baselines' respective candidate substitute nomination methods and conduct one-to-one comparison.

\subsubsection*{Decision-based Baseline Methods}
As stated in related work, to the best of our knowledge, there are only a few decision-based attack models and none of them are word-level.
Therefore, we choose a sentence-level decision-based attack model \citep{zhao2018generating} as the baseline. 
This model uses an autoencoder to encode the original sentence into a vector, then adds some noise and finally decodes the perturbed vector into a sentence, i.e., a potential adversarial example.
This process is repeated until an adversarial example is generated that successfully fools the victim model.
We denote this model as \textbf{AED}.


Furthermore, to make the comparison more extensive, we adapt two score-based baselines including GA+Embedding and PSO+Sememe to the decision-based attack setting.\footnote{PWWS+Synonym cannot be adapted to the decision-based attack setting.}
They originally use the variation of the prediction scores given by the victim model to direct their attacks.
We simply change the required prediction score variation to a binary success/fail signal.
We denote the two revised methods as \textbf{GA'+Embedding} and \textbf{PSO'+Sememe} respectively.

\begin{figure*}[!t]
    \centering
    {
    \includegraphics[width=4.8cm]{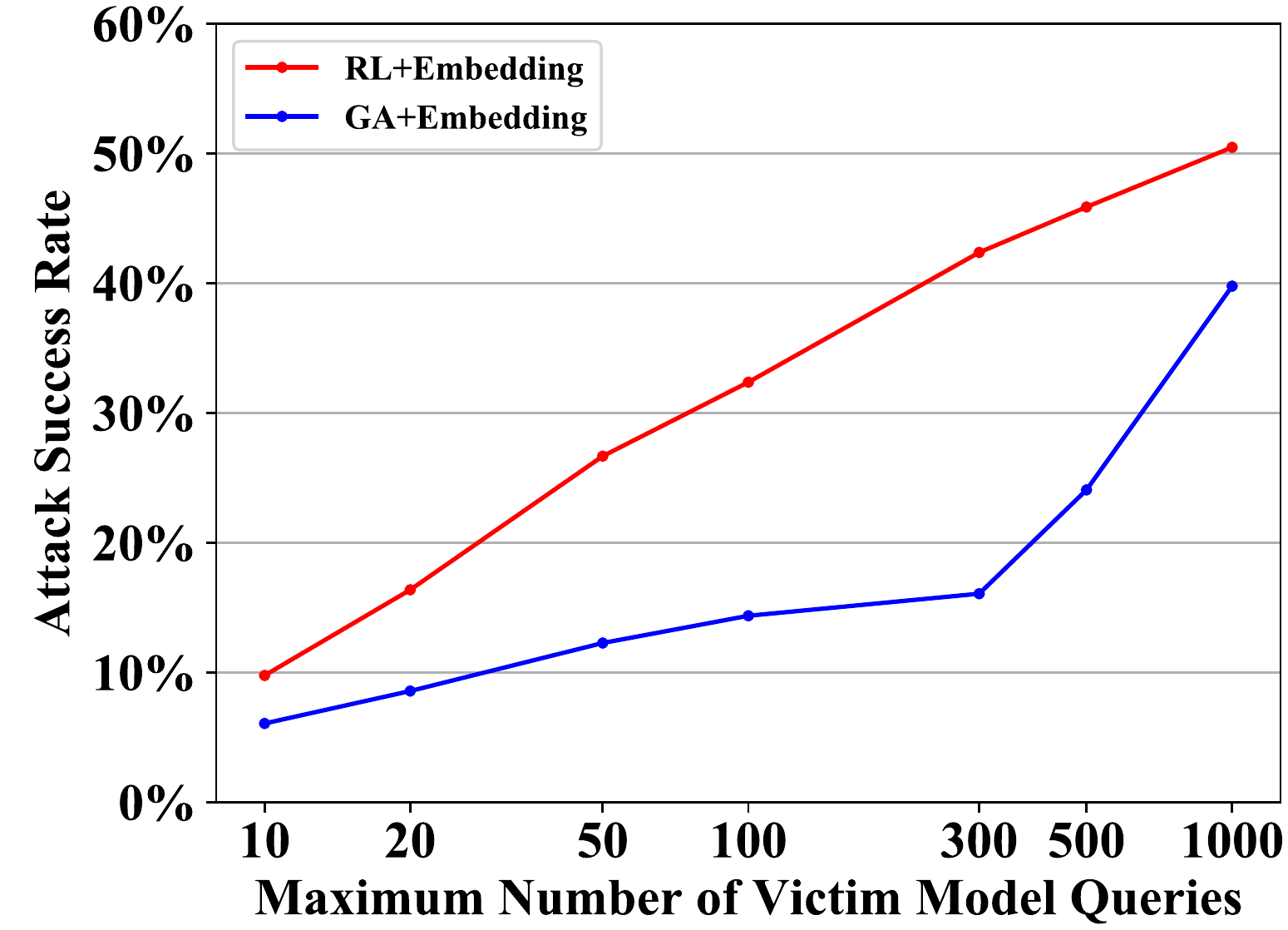}
    }
    \hspace{0in}
    {
    \includegraphics[width=4.8cm]{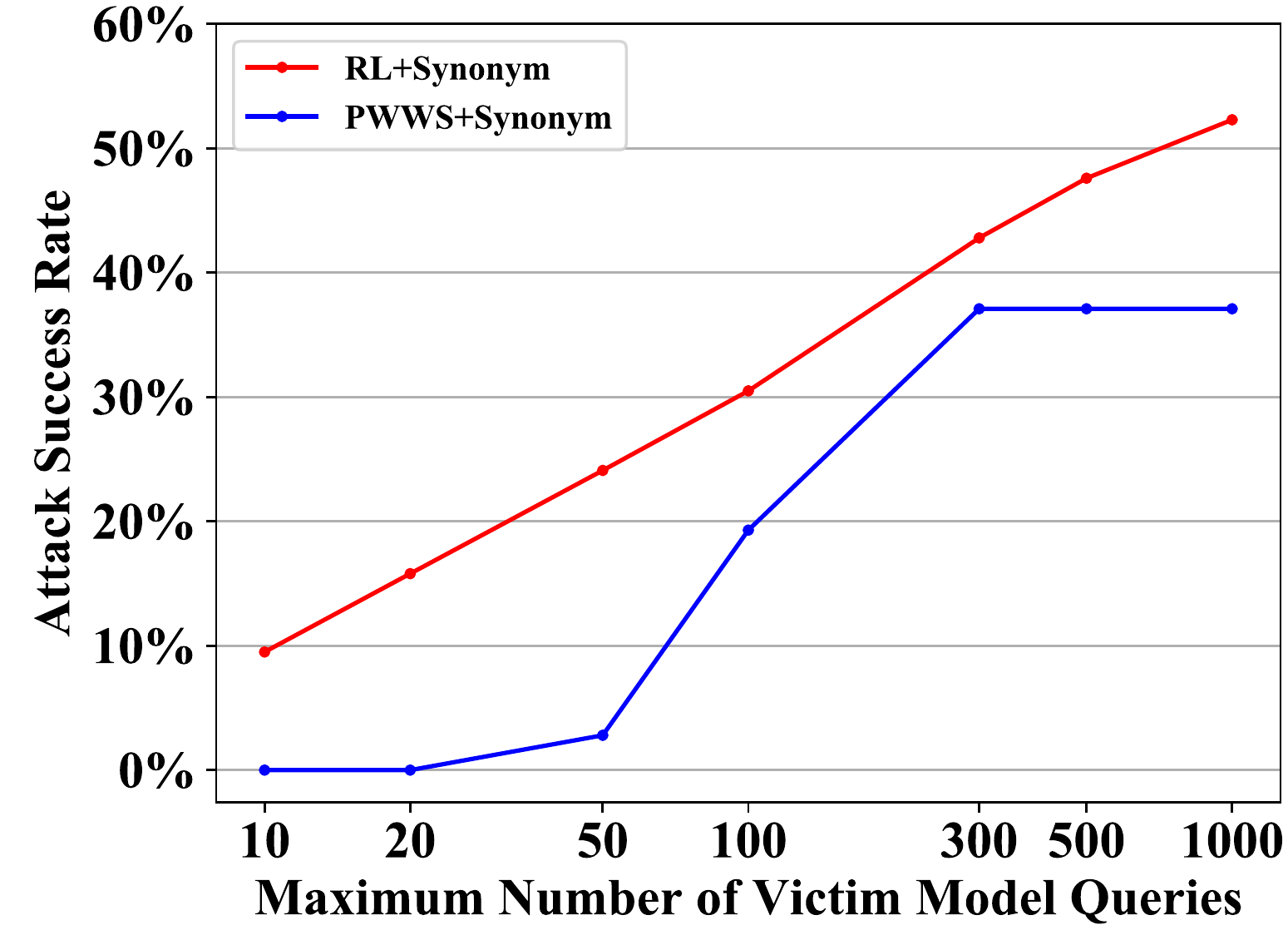}
    }
    \hspace{0in}
    {
    \includegraphics[width=4.8cm]{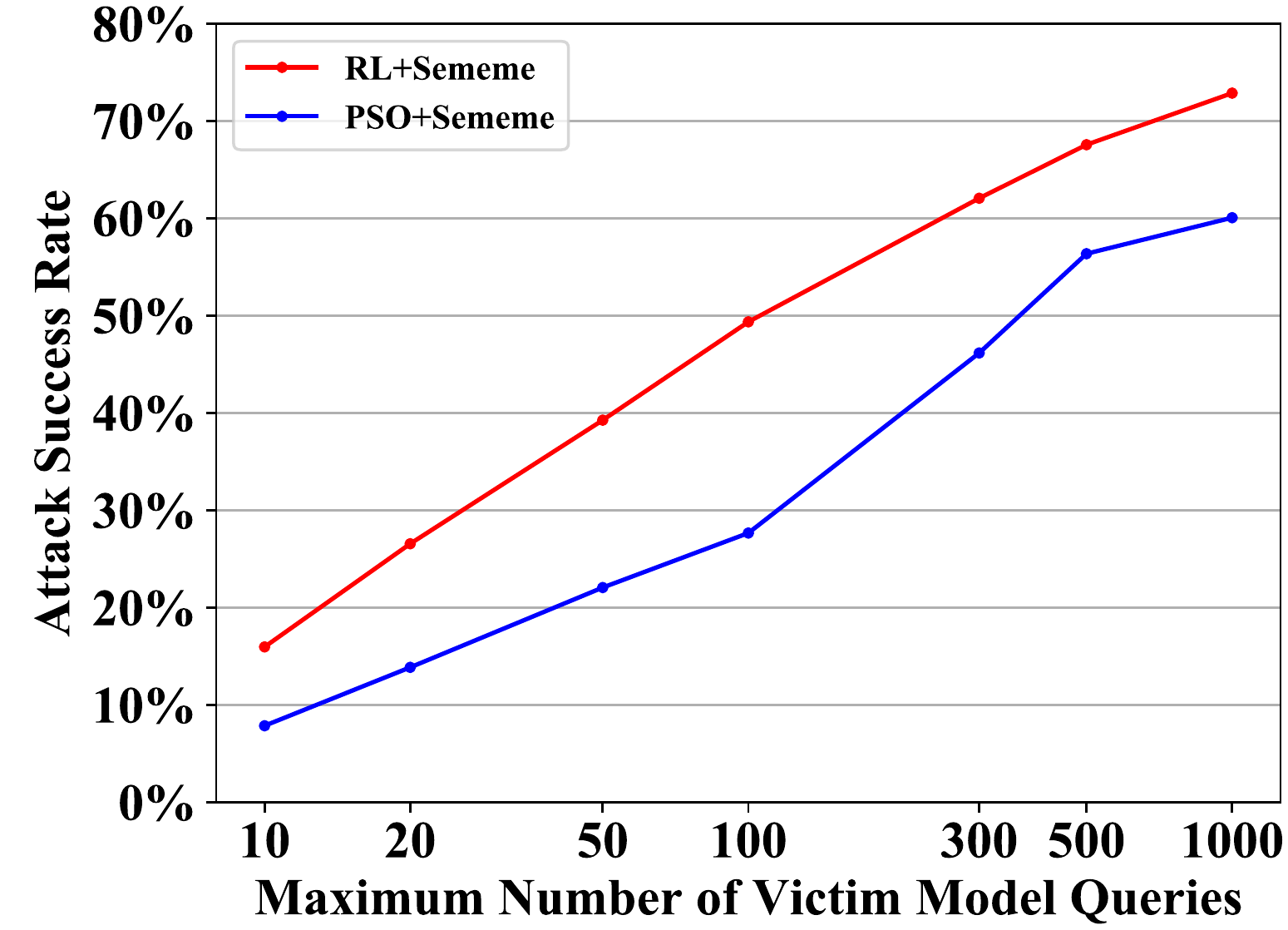}
    }
    \caption{Attack success rates of different score-based attack models against ALBERT on SST-2.
    }
\label{fig:score-SST}
\end{figure*}

\begin{figure*}[!t]
    \centering
    {
    \includegraphics[width=4.8cm]{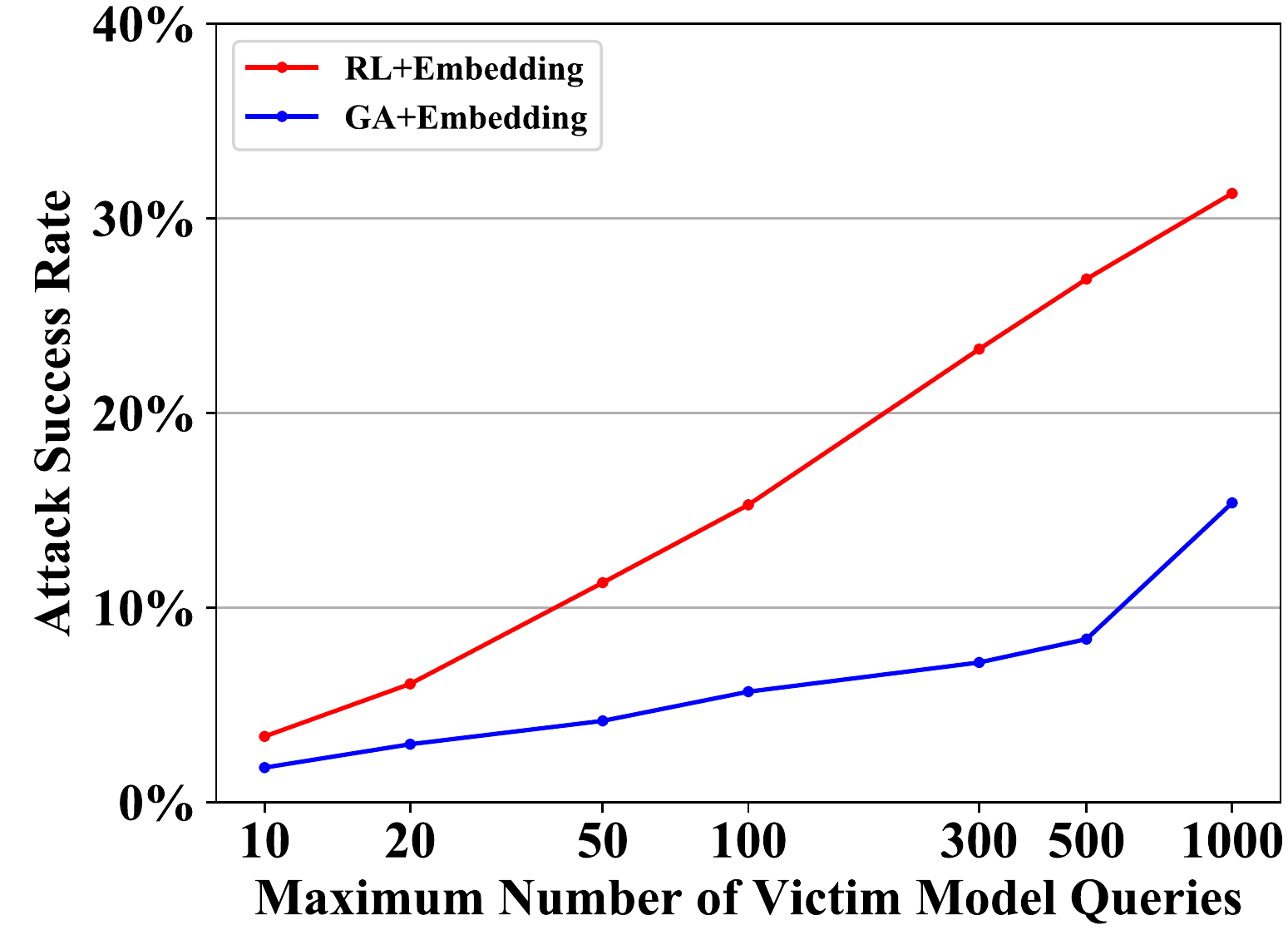}
    }
    \hspace{0in}
    {
    \includegraphics[width=4.8cm]{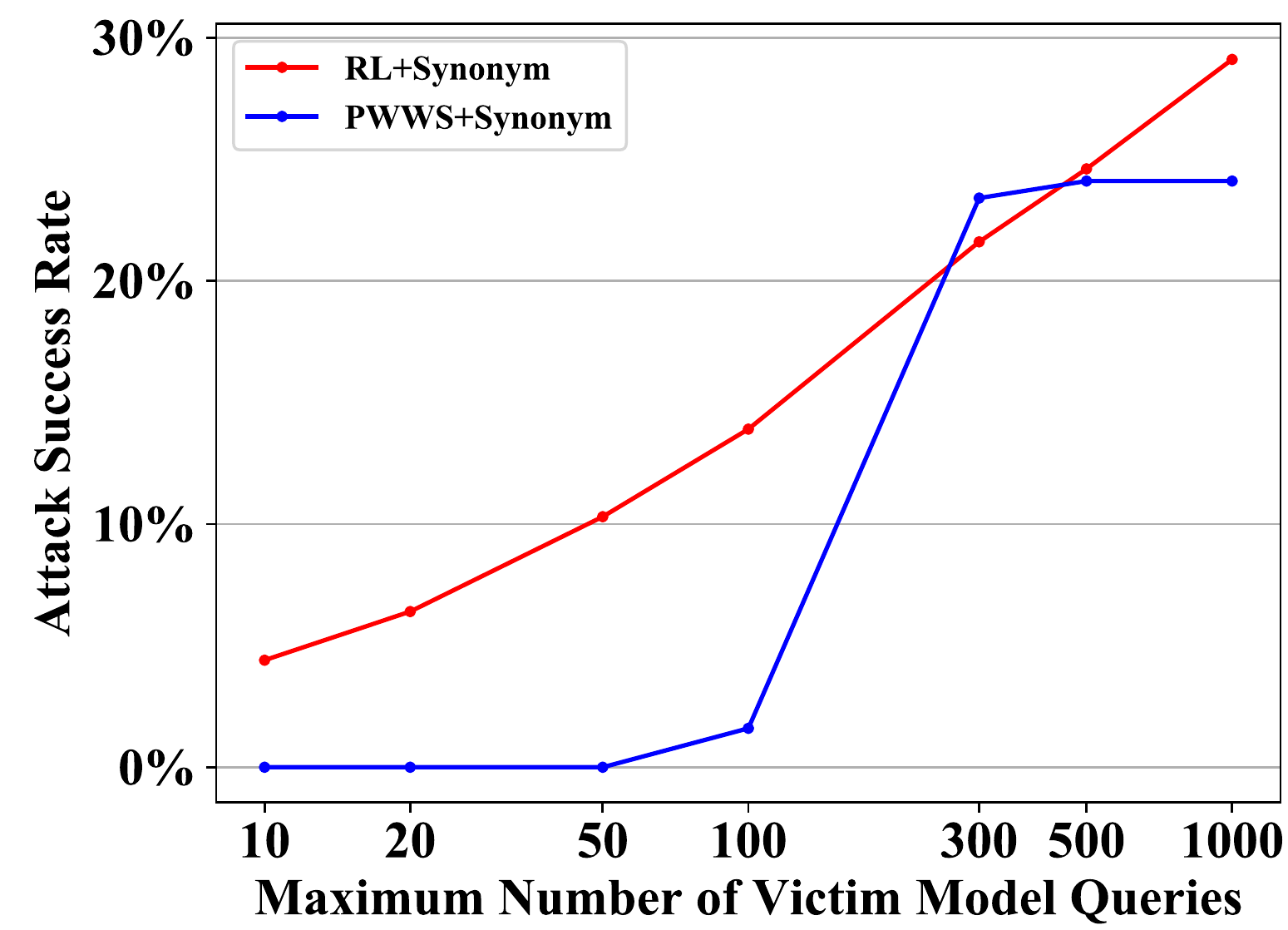}
    }
    \hspace{0in}
    {
    \includegraphics[width=4.8cm]{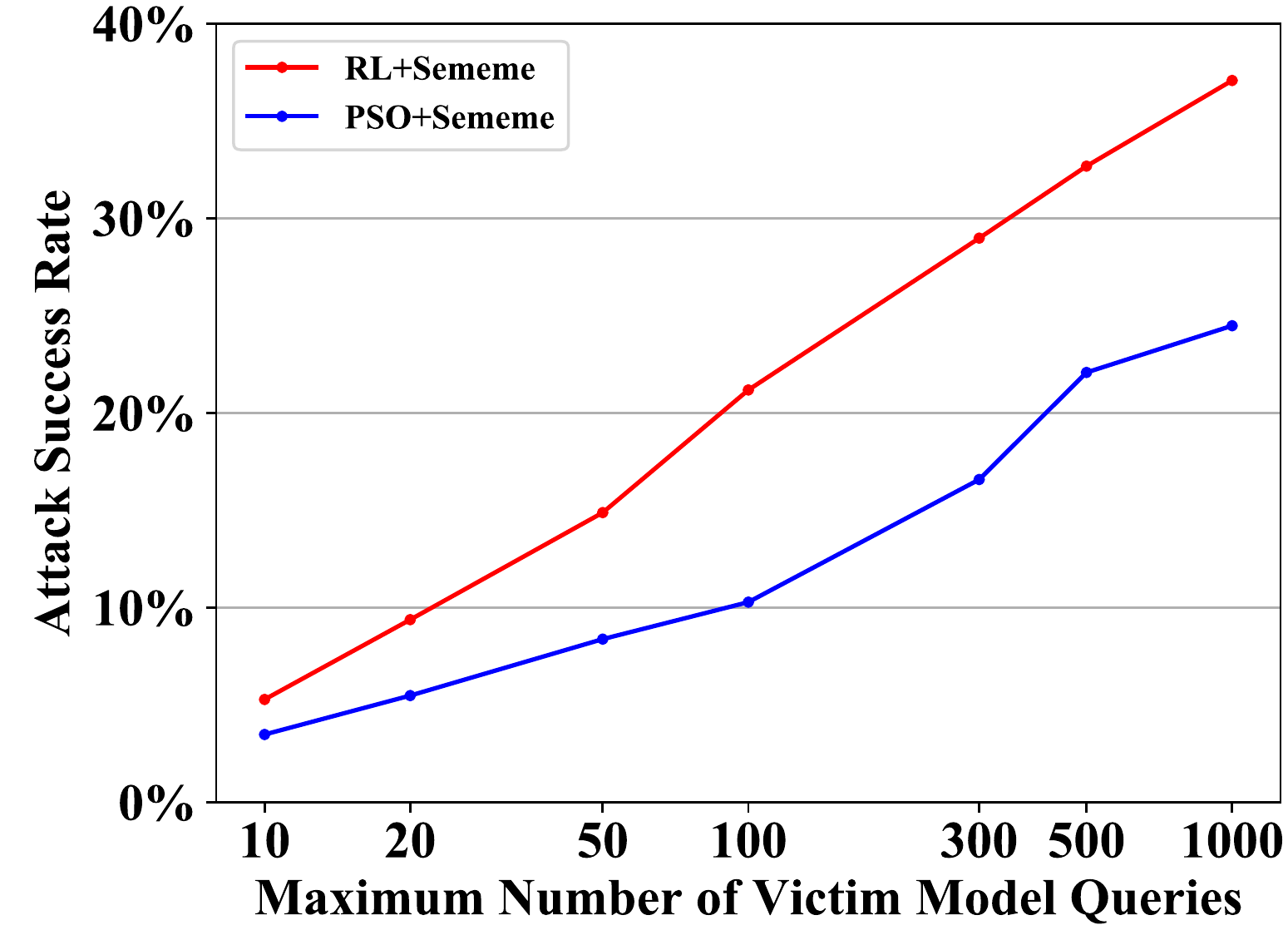}
    }
    \caption{Attack success rates of different score-based attack models against XLNet on AG News.}
\label{fig:score-AG}
\end{figure*}

\begin{figure*}[!t]
    \centering
    {
    \includegraphics[width=4.8cm]{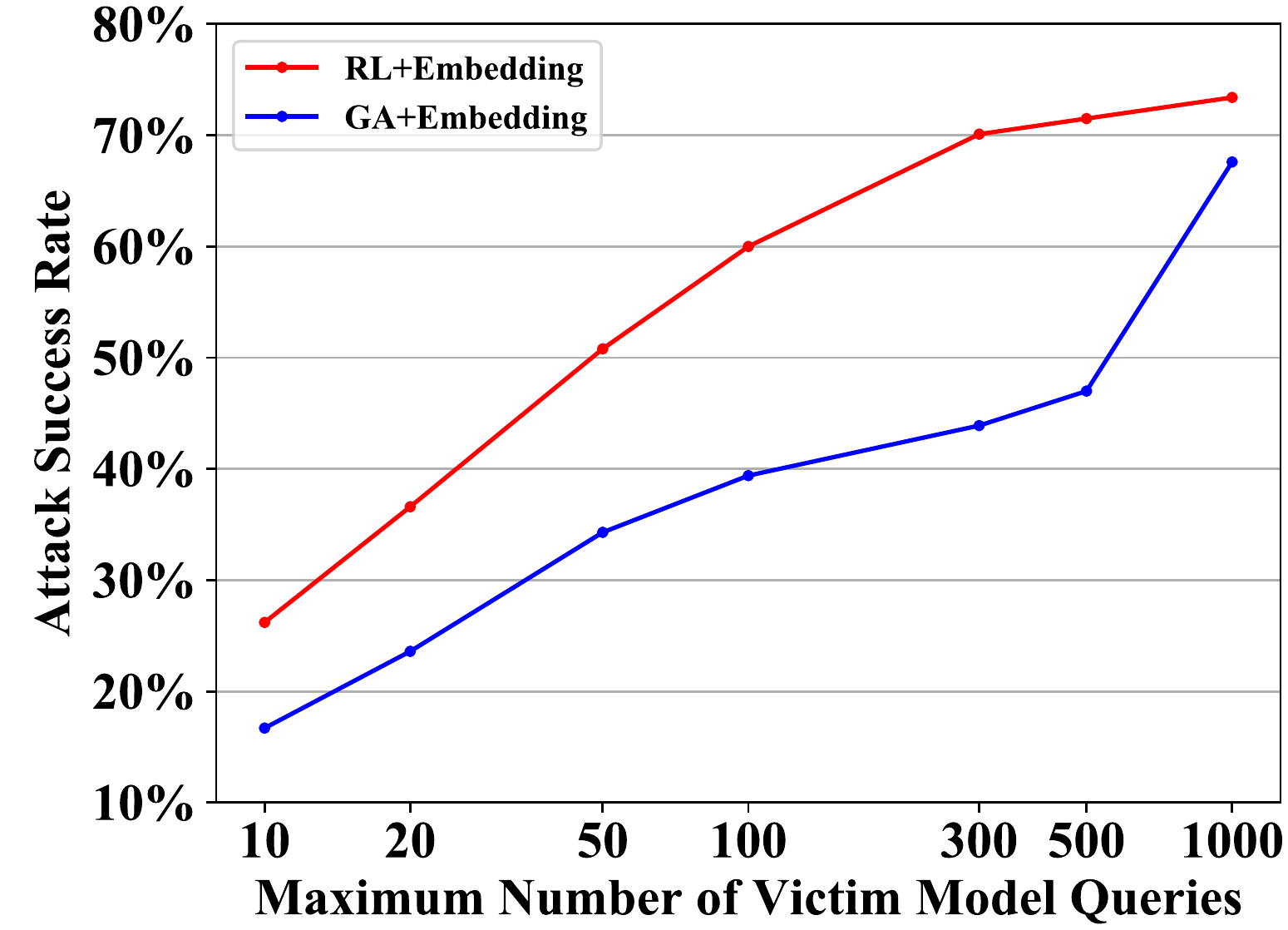}
    }
    \hspace{0in}
    {
    \includegraphics[width=4.8cm]{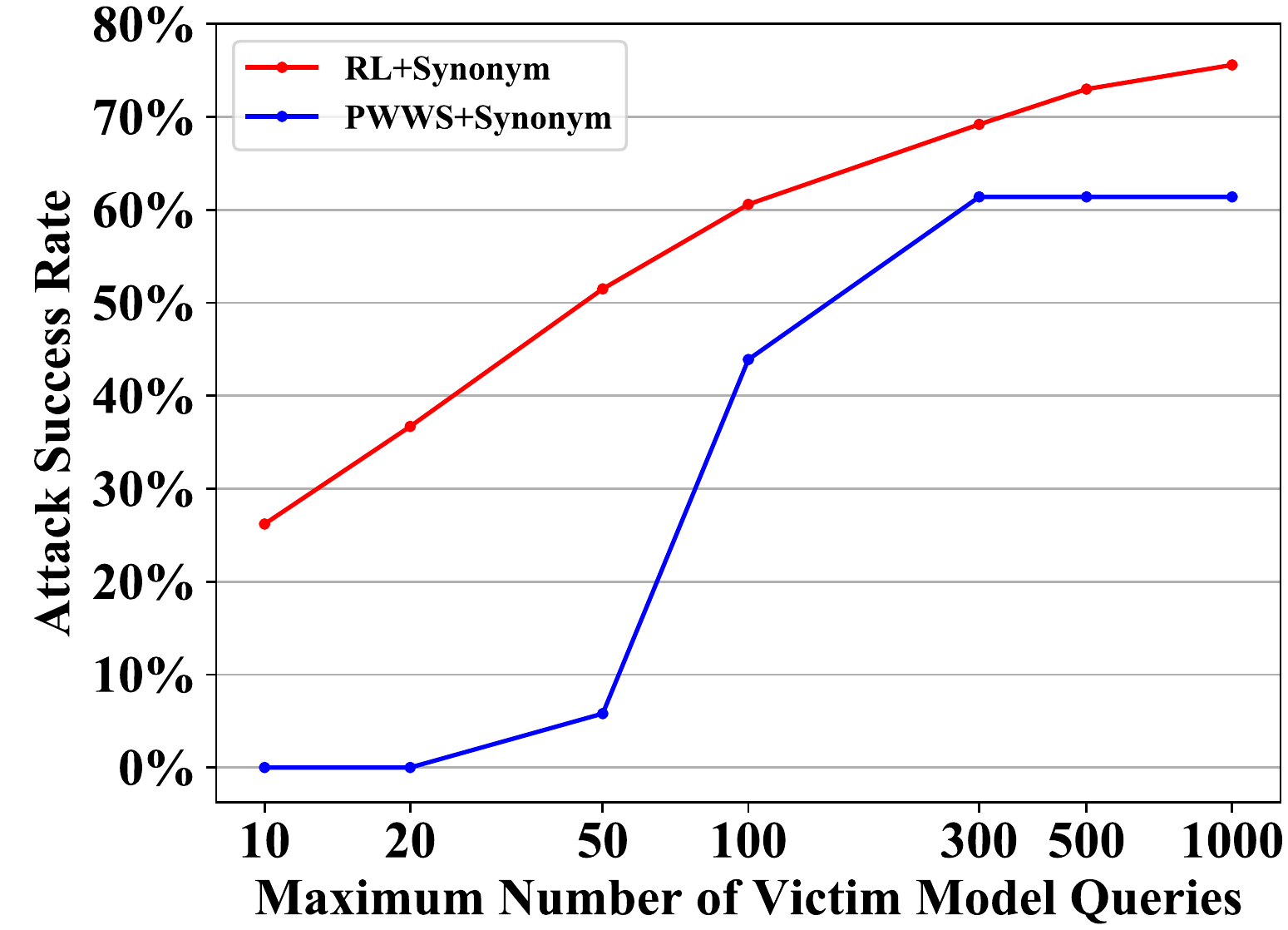}
    }
    \hspace{0in}
    {
    \includegraphics[width=4.8cm]{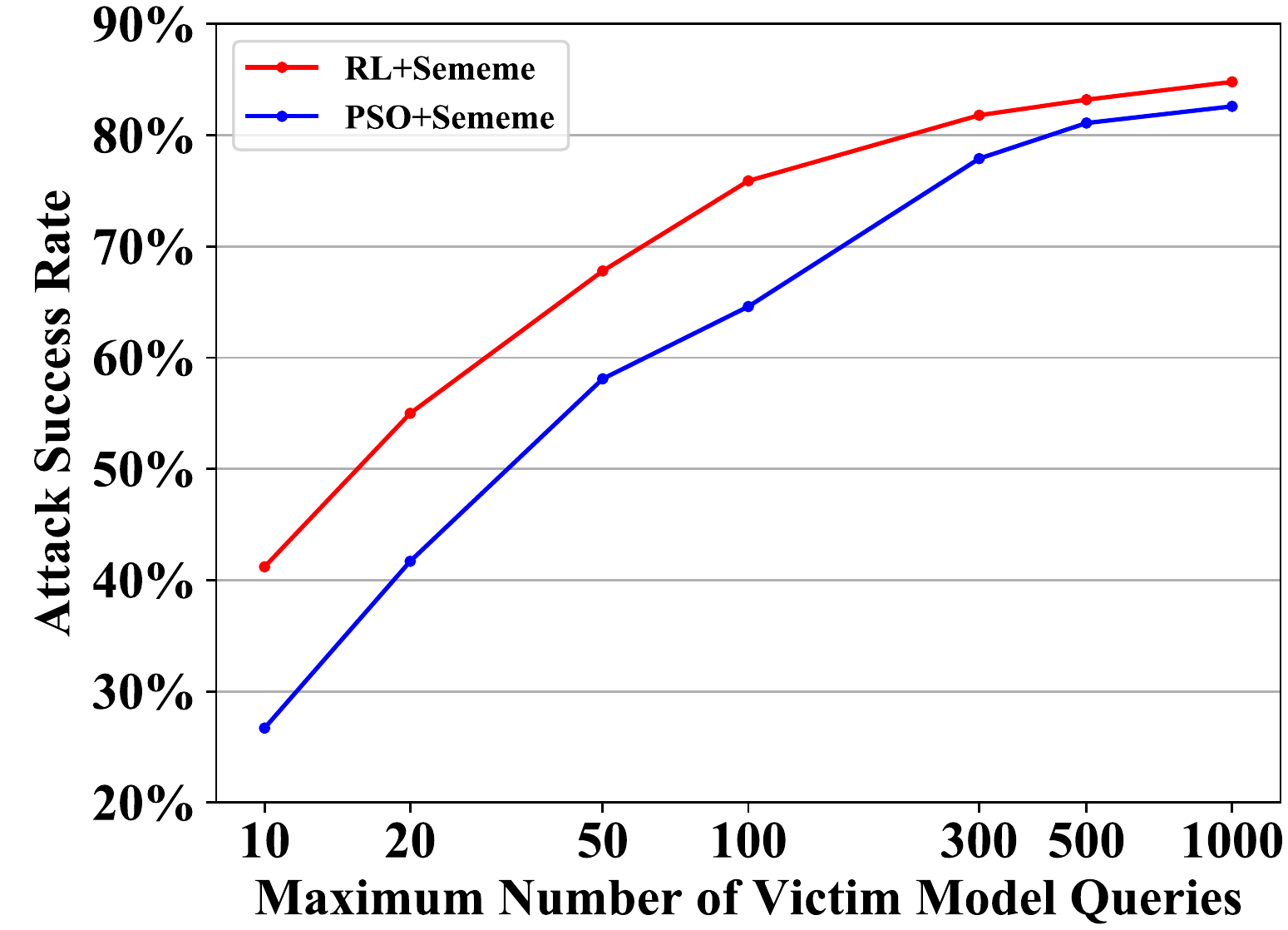}
    }
    \caption{Attack success rates of different score-based attack models against RoBERTa on MNLI-m.}
\label{fig:score-MNLI}
\end{figure*}

\begin{figure*}[!t]
    \centering
    {
    \includegraphics[width=4.8cm]{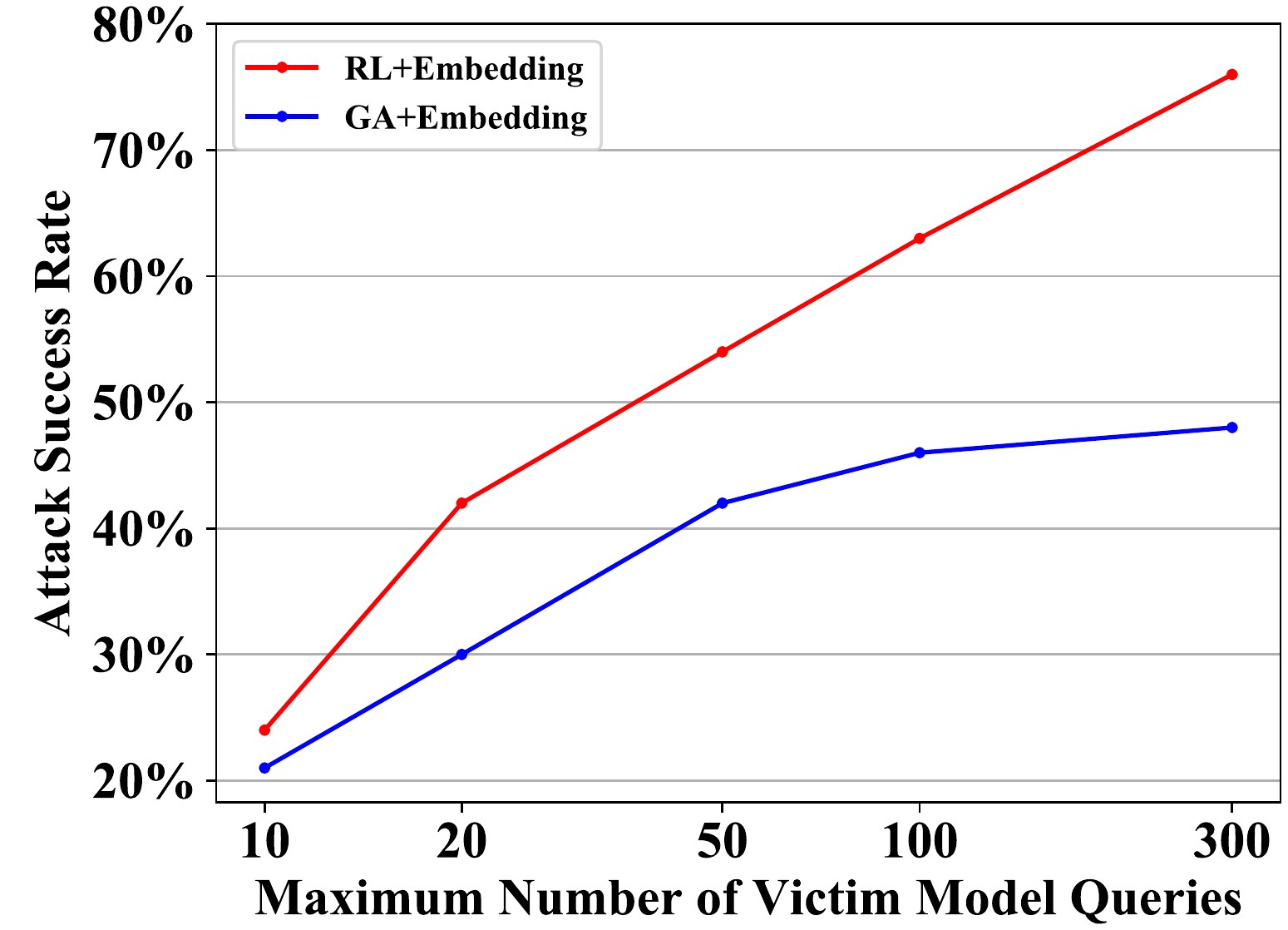}
    }
    \hspace{0in}
    {
    \includegraphics[width=4.8cm]{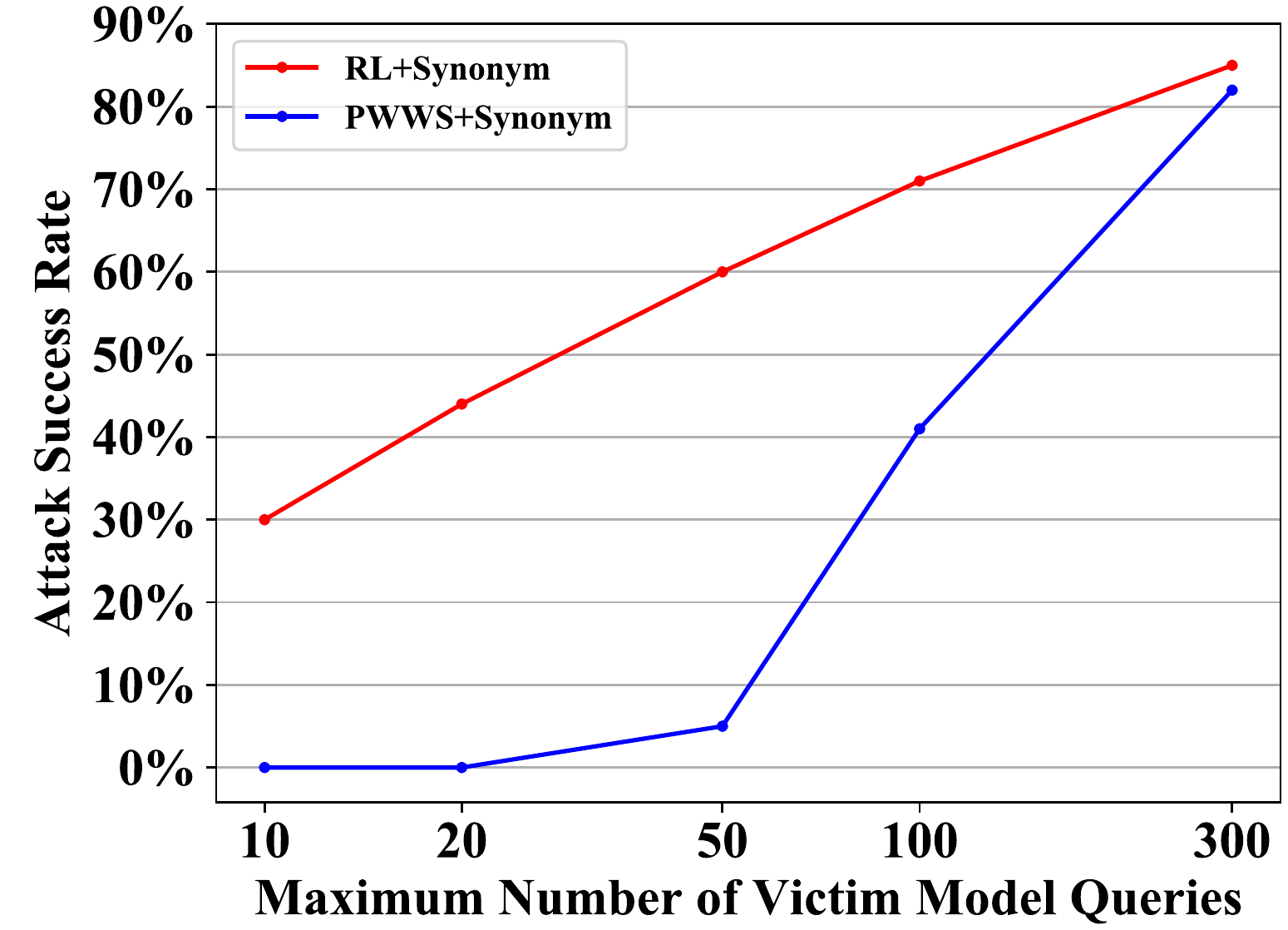}
    }
    \hspace{0in}
    {
    \includegraphics[width=4.8cm]{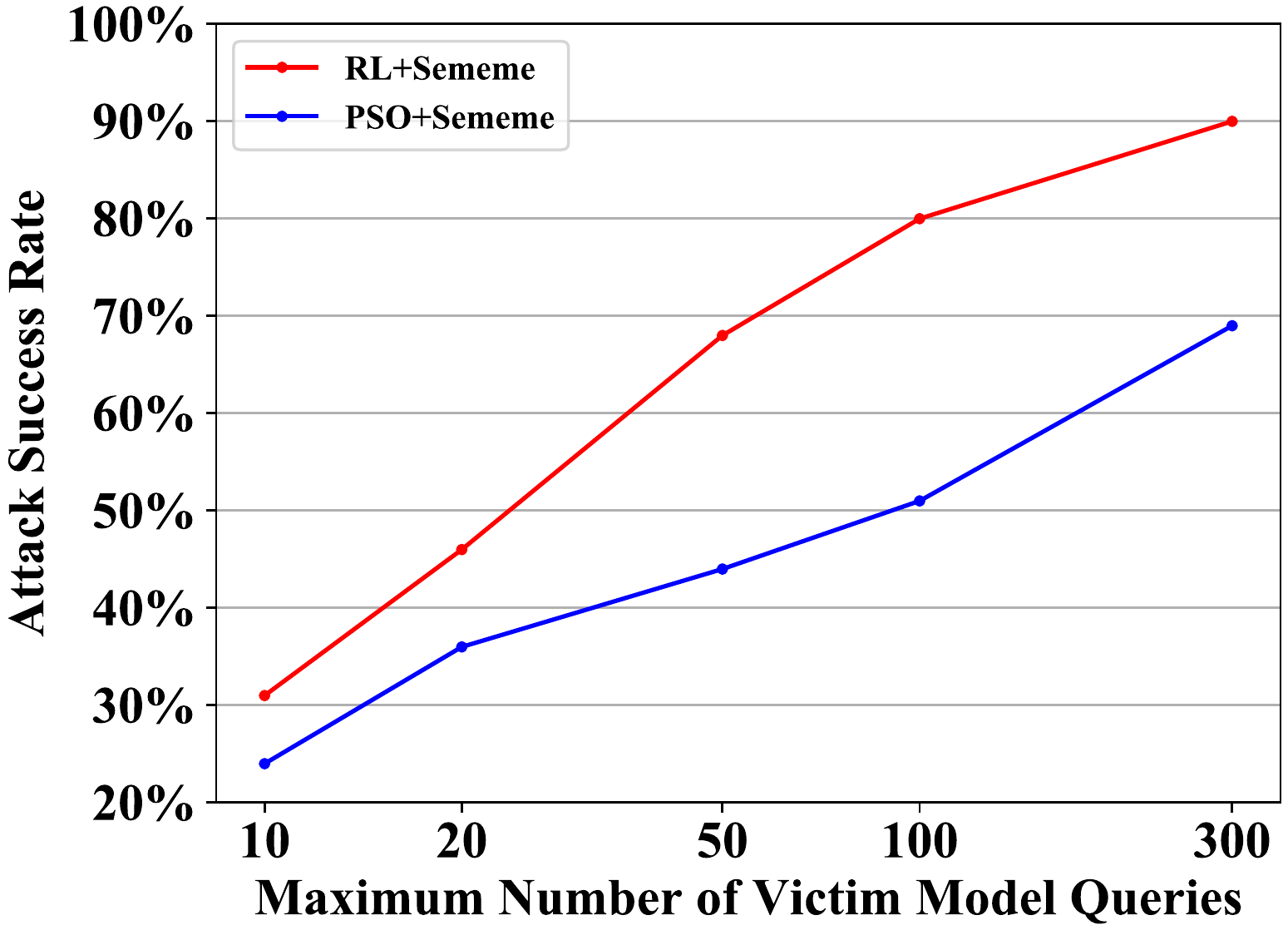}
    }
    \caption{Attack success rates of different score-based attack models against Microsoft Azure API on SST-2. Considering the time (about 1s per query) and cost (about \$0.2 per 100 queries) of accessing the API, the upper limit of the maximum number of victim queries is $300$ rather than $1,000$ as for the other victim models. The same is true for the Meaning Cloud API in the decision-based attack setting.
    }
\label{fig:score-API}
\vspace{-0.8em}
\end{figure*}

\begin{figure*}[!tb]
    \centering
    \subfigure[Attacking ALBERT on SST-2]
    {          
        \includegraphics[width=.48\linewidth]{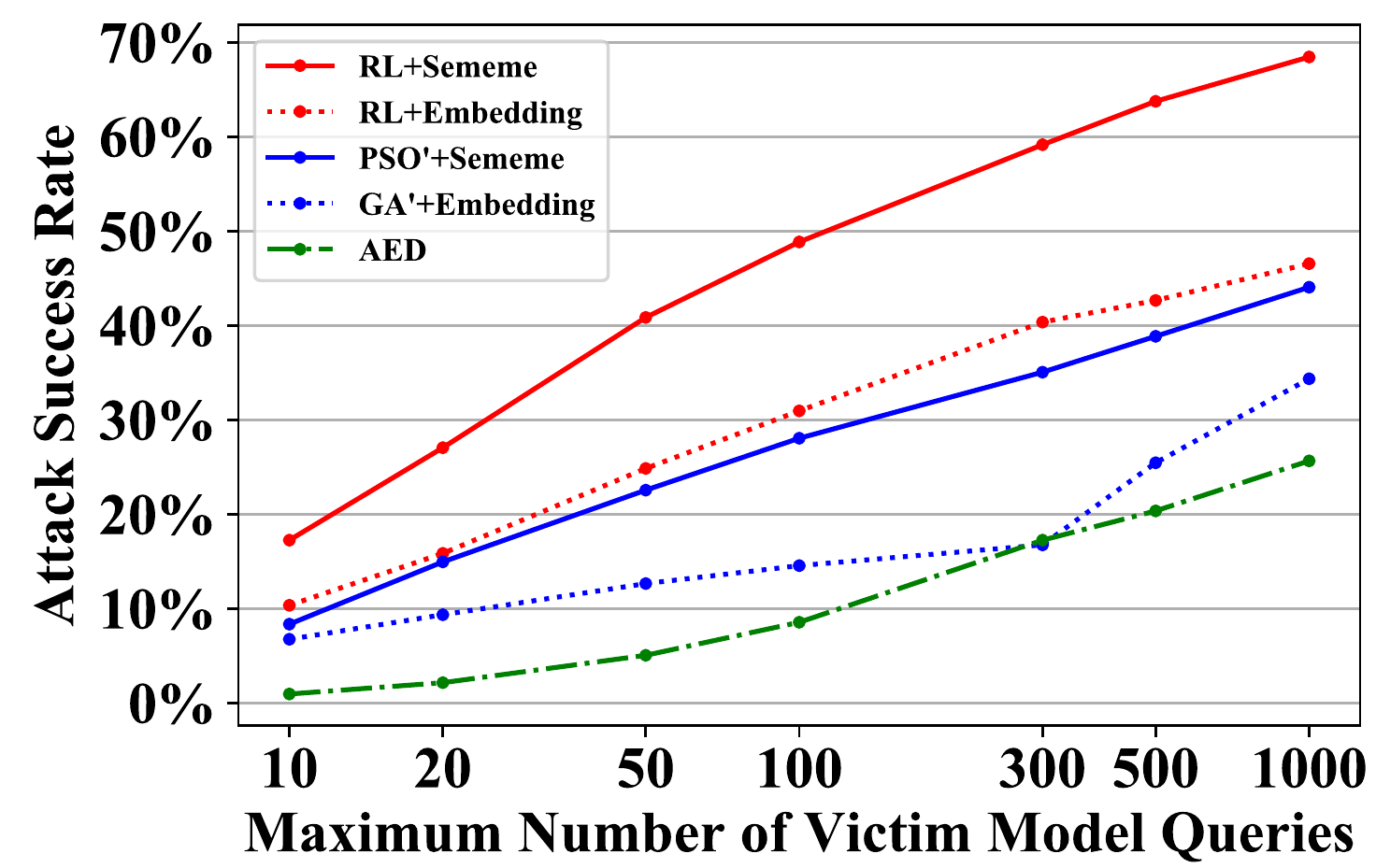}
    }
    \hspace{0cm}
    \subfigure[Attacking XLNet on AG News]
    {
        \includegraphics[width=.48\linewidth]{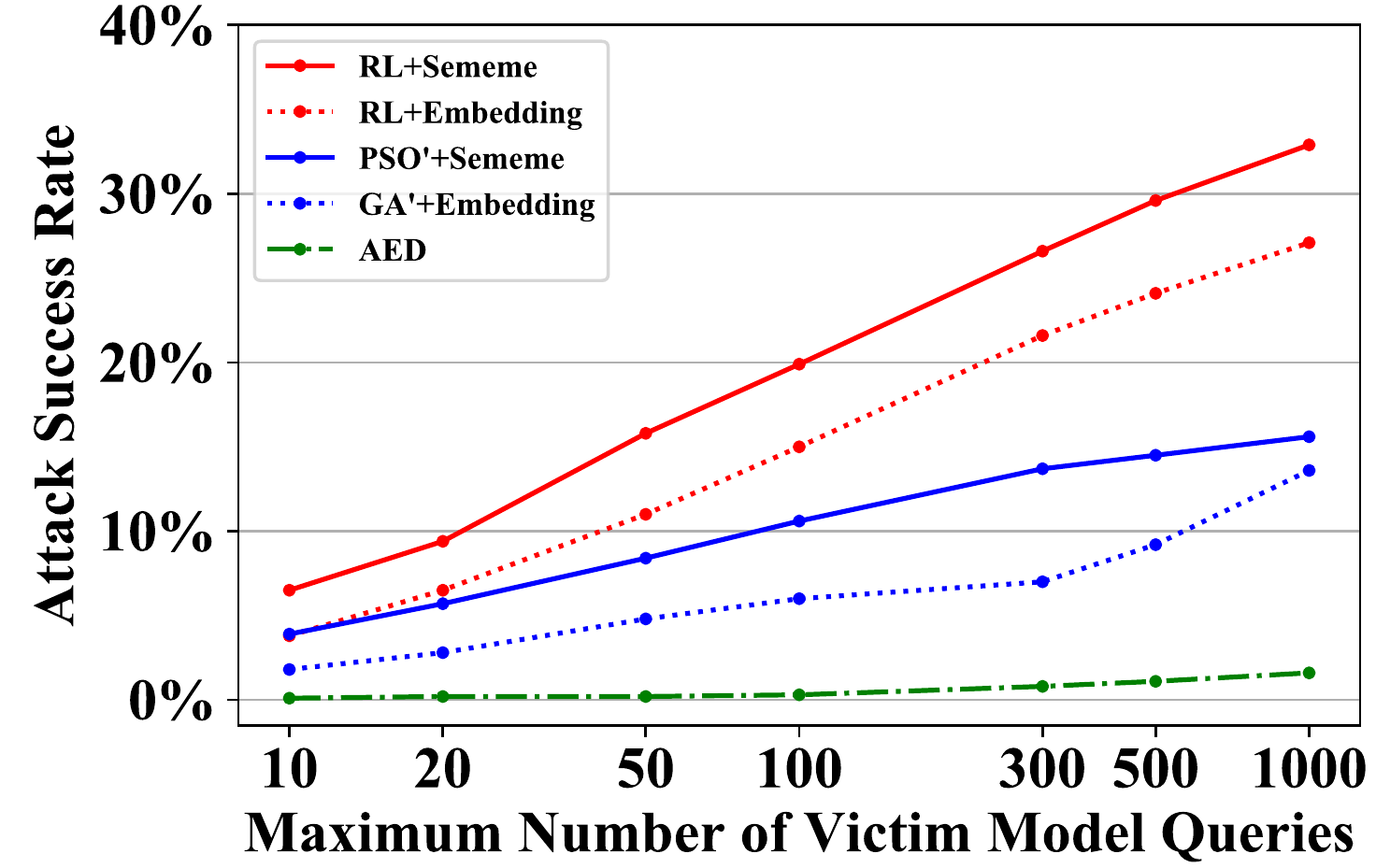}
    }
    \subfigure[Attacking RoBERTa on MNLI-m]{              
    \includegraphics[width=.48\linewidth]{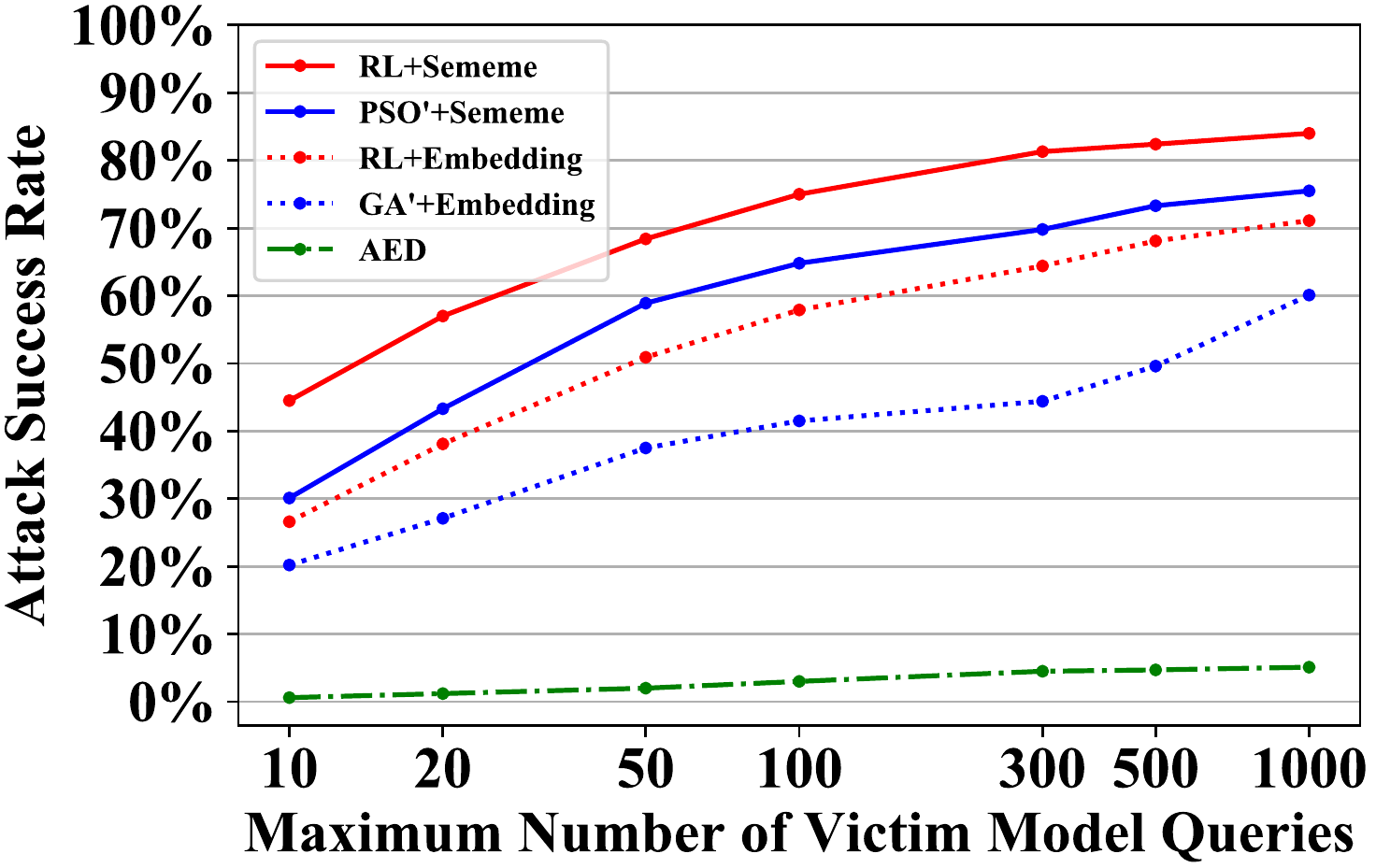}}
    \hspace{0cm}
    \subfigure[Attacking Meaning Cloud API on SST-2]{
    \includegraphics[width=.48\linewidth]{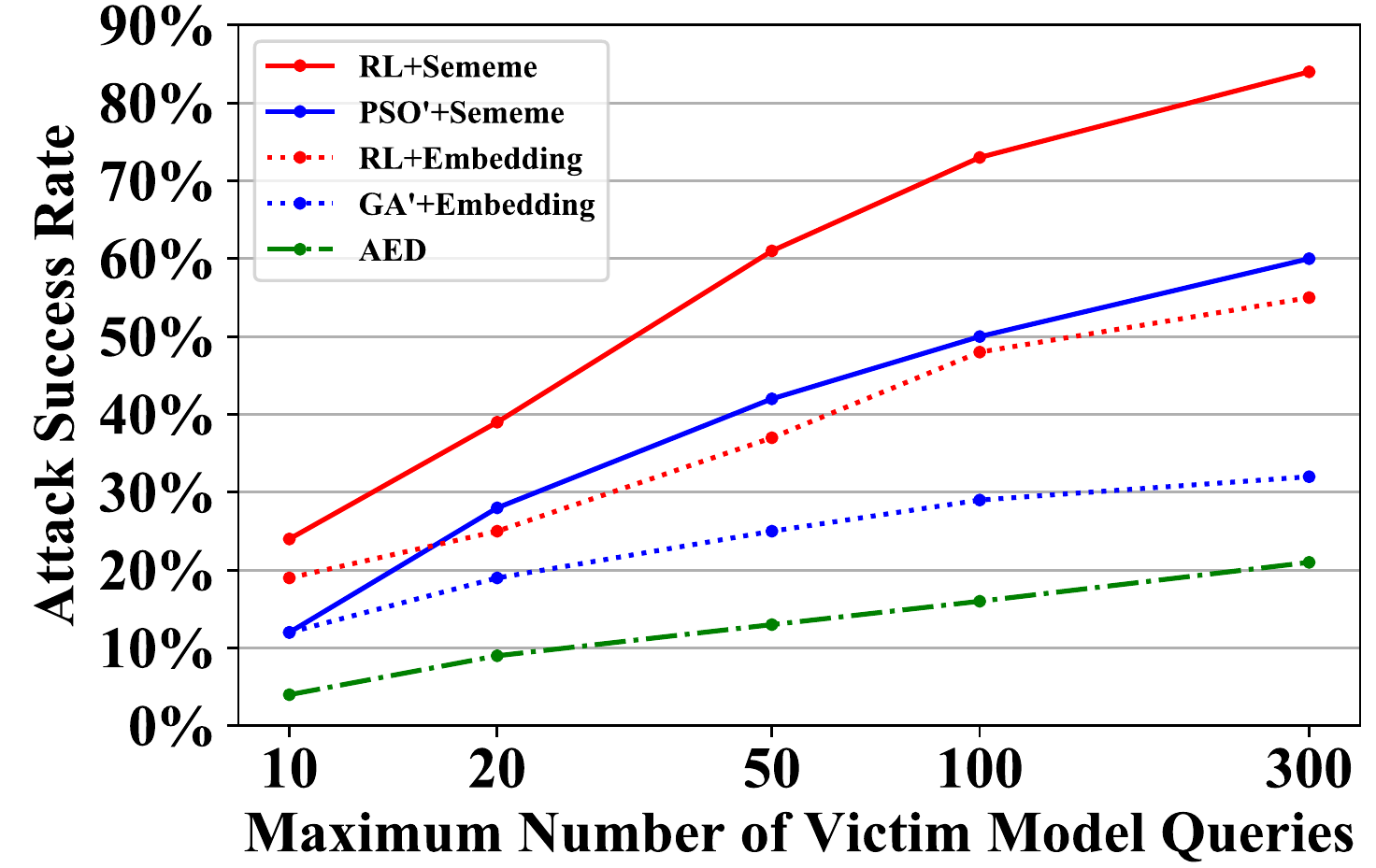}}
    \caption{Attack success rates of different decision-based attack models.}
\label{fig:decision}
\vspace{-0.5em}
\end{figure*}

\subsection{Experimental Settings}
\paragraph{Hyper-parameters and Training}
The learning rates of $\mathbf{p}^x$ and $\mathbf{q}^x_i$ in our attack model are set to $0.2$ and $0.5$ respectively.
The discount factor $\gamma$ is $0.4$.
In the decision-based attack setting, the multi-layer perceptron of the pre-trained score-based attack model has two layers whose sizes are $768\times32$ and $32\times1$ respectively, the virtual victim model is BERT, and the the reward is $-1$.
More details of our model's hyper-parameter settings are given in the appendix.
For all the baseline methods, we use their recommended hyper-parameter settings.

\paragraph{Evaluation Settings}
Following previous work \citep{alzantot2018generating,zang2020word}, we restrict the length of attacked original sentences to 10-100, and set the maximum word modification rate of adversarial examples to 25\%.
We evaluate the attack performance from three perspectives including attack success rate, attack efficiency and attack validity.
(1) Attack success rate is the percentage of the attacks that successfully craft adversarial examples and fool the victim model.
We calculate the attack success rates of attack models within different limits on the maximum number of victim model queries.
(2) To evaluate attack efficiency, we use the average number of victim model queries used for attacking an instance as the metric.
(3) Attack validity reflects whether an adversarial example has the same ground-truth label as the original sentence, and we carry out human evaluation to evaluate it.
Details of human evaluation are given in the appendix.





\begin{table}[t!]
\resizebox{\linewidth}{!}{
\begin{tabular}{c||c|c|c|c}
\toprule
\multirow{2}{*}{Attack Model} & SST-2 & AG News & MNLI-m & SST-2 \\
\cline{2-5}
& ALBERT          & XLNet          & RoBERTa           & API \\ 
\hline
GA+Embedding                  & 365.69          & 639.68        & 228.44        & \ \ 27.65                            \\  
RL+Embedding                  & \textbf{\ \ 83.96} & \textbf{\ \ 77.69} & \textbf{\ \ 48.14} & \textbf{\ \ 17.83}                     \\ 
\hline
PWWS+Synonym                  & 101.15                   & 173.03                     &  \ \ 85.61                              & 100.06                            \\ 
RL+Synonym                    & \textbf{\ \ 92.51}                              & \textbf{153.17}                              & \textbf{\ \ 47.49}                     & \textbf{\ \ 58.42}                     \\ 
\hline
PSO+Sememe                    & 177.85                              & 212.24                              & \ \ 71.81                              & \ \ 65.22                             \\ 
RL+Sememe                     & \textbf{\ \ 80.85}                     & \textbf{\ \ 92.81}                     & \textbf{\ \ 38.64}                     & \textbf{\ \ 53.54}  \\       
\bottomrule          
\end{tabular}
}
\caption{Average numbers of victim model queries in the \textit{score}-based attack setting.
}
\label{tab:score_query}
\vspace{-0.5em}
\end{table}

\begin{table}[t!]
\resizebox{\linewidth}{!}{
\begin{tabular}{c||c|c|c|c}
\toprule
\multirow{2}{*}{Attack Model} & SST-2 & AG News & MNLI-m & SST-2 \\
\cline{2-5}
& ALBERT          & XLNet          & RoBERTa           & API \\ 
\hline
AED & 260.78 &  345.00 & 140.29 & 249.60 \\ \hline
GA'+Embedding                  & 299.14         & 557.07        & 162.52       & \textbf{\ \ 29.48}                            \\  
RL+Embedding                  & \textbf{\ \ 69.79} & \textbf{\ \ 55.00} & \textbf{\ \ 49.62} & \ \ 33.90                     \\ \hline
PSO'+Sememe                   & 158.72                              & 117.49                              & \ \ 66.33                             & \ \ 49.45                             \\ 
RL+Sememe                     & \textbf{\ \ 51.42}                     & \textbf{\ \ 56.56}                     & \textbf{\ \ 27.58}                     & \textbf{\ \ 33.24}  \\       \bottomrule          
\end{tabular}
}
\caption{Average numbers of victim model queries in the \textit{decision}-based attack setting.
}
\label{tab:decision_query}
\vspace{-1em}
\end{table}

\subsection{Experimental Results}
\paragraph{Attack Success Rate}
Figure \ref{fig:score-SST}-\ref{fig:score-API} illustrate the attack success rates of our attack model and three baseline methods against four victim models (including an API) in the score-based attack setting respectively, and Figure \ref{fig:decision} shows the attack success rates in the decision-based attack setting.
We can observe that in both attack settings, our model consistently achieves higher success rates than all the baseline methods against whichever victim model, which demonstrates the superiority of our model.
In addition, we find that the limit of victim model queries has considerable impact on attack success rate.
When maximum number of victim model queries is very small, the attack success rates are quite low, which reflects the importance of attack efficiency in realistic attack situations.

\paragraph{Attack Efficiency}
Table \ref{tab:score_query} and \ref{tab:decision_query} list the average numbers of victim model queries of different attack models against different victim models in the score- and decision-based attack settings, respectively.\footnote{These results are based on the setting that the maximum number of victim model queries is $1000$ for ALBERT, XLNet and RoBERTa, and $300$ for the two APIs.}
We observe that our model basically needs least queries of the victim model, even though it has the highest attack success rates (as shown in Figure \ref{fig:score-SST}-\ref{fig:decision}).
It demonstrates the high efficiency of our model.

\vspace{-0.3em}
\paragraph{Attack Validity}
According to the results of human evaluation and significance tests, our attack model performs equally as compared with baseline methods in terms of attack validity.
More details of the attack validity results are given in the appendix.

\begin{figure}[!tb]
    \centering
    \includegraphics[width = .95\linewidth]{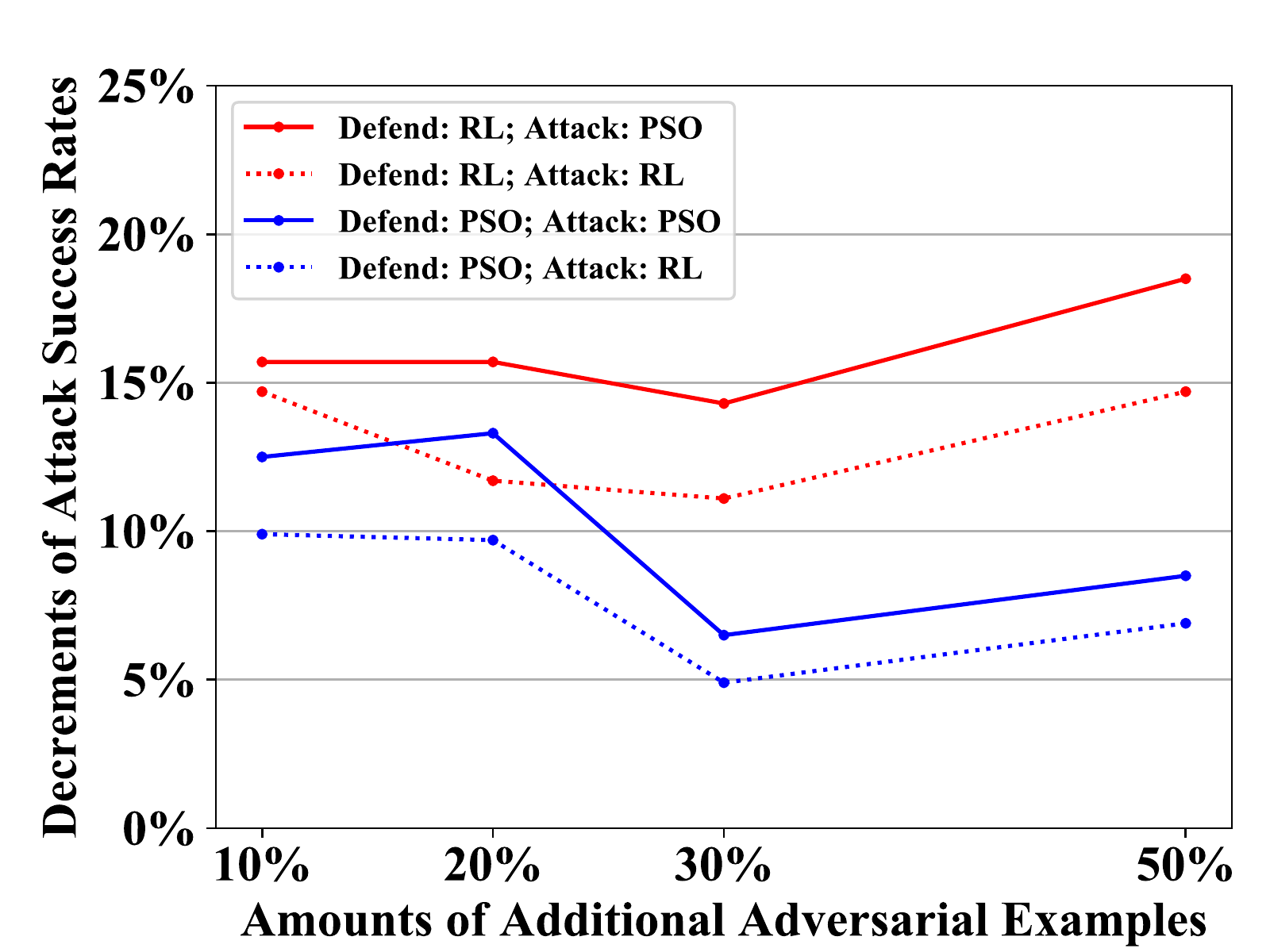}
    \caption{Decrements of attack success rates brought by adversarial training using different amounts of adversarial examples (denoted by the ratio to training size).
    ``Defend'' represents the attack model used for adversarial training.
    }
    \label{fig:adv_training}
\vspace{-1.2em}
\end{figure}

\subsection{Adversarial Training}
Adversarial training, which augments training data with adversarial examples, is believed to be an effective method of improving model robustness against adversarial attacks \citep{goodfellow2015explaining}.
In this experiment, we try re-training ALBERT using SST-2's training data augmented with different amounts of adversarial examples that are generated by attacking the training instances, and observe the change of attack success rates under score-based adversarial attacks.
We conduct comparison with PSO+Sememe, which has the highest attack performance among three baseline methods.

Figure \ref{fig:adv_training} shows the results, where the maximum number of victim model queries is $1,000$.
From the perspective of defense, by comparing the results from different adversarial training methods, namely \textcolor{red}{---} vs. \textcolor{blue}{---} as well as  \textcolor{red}{$\cdots$} vs. \textcolor{blue}{$\cdots$}, we can find that the adversarial examples generated by our model can bring more robustness improvement to the victim model than PSO, even against PSO's attacks.
From the perspective of attack, by comparing the results from different attack models, namely 
\textcolor{red}{---} vs. \textcolor{red}{$\cdots$} as well as \textcolor{blue}{---}  vs. \textcolor{blue}{$\cdots$}, we conclude that our model is more difficult to overcome by adversarial training.

\section{Conclusion and Future Work}
In this paper, we propose a reinforcement learning-based textual adversarial attack model aimed at real-world adversarial attack situations.
It can work in both score- and decision-based attack settings and possesses learning ability so as to launch attacks more efficiently. 
We also find that our model can bring more robustness improvement to the victim model by adversarial training as compared with existing baselines.

In the future, we will work towards further enhancing attack efficiency and improving attack performance in the situation of extremely limited victim model queries.
In addition, we will explore how to make model more robust by adversarial training or other methods.

\bibliographystyle{acl_natbib}
\bibliography{emnlp2020}

\end{document}


\maketitle

\appendix


\section{Datasets and Preprocessing}
Each of the three datasets is split into train, validation, and test sets. 
Original SST-2 dataset\footnote{\url{https://nlp.stanford.edu/sentiment/index.html}} has prepared the three sets so we directly use the original version. 
For AG New\footnote{\url{http://groups.di.unipi.it/~gulli/AG_corpus_of_news_articles.html}}, we randomly select $24,000$ instances from its training set
as the validation set. 
Since the test set of MNLI\footnote{\url{https://cims.nyu.edu/~sbowman/multinli/}} is not labeled, we use its original validation set as the test set, and randomly select $9,817$ instances from original training set as the validation set.
We use Keras text preprocessing API to tokenize the corpus. 
And we use the Stanford Tagger\footnote{\url{https://nlp.stanford.edu/software/tagger.shtml}} to conduct part-of-speech tagging, which is required by some candidate substitute nomination methods.

\section{Hyper-parameter Searching}
We manually tune the hyper-parameters of our attack models and choose the best on the validation set. 
For simplicity, we only tune the hyper-parameters on {SST-2} and apply the tuned hyper-parameters to the other datasets.

For the learning rate of $\mathbf{p}^x$ and $\mathbf{q}^x_i$ in our model, we tune them in the range of $0.1$ to $0.7$ for about 10 trials and finally set the learning rate of $\mathbf{p}^x$ to $0.2$ and that of $\mathbf{q}^x_i$ to $0.5$.
For the discount factor $\gamma$ we tune it in the range of $0.2$ to $0.6$ for about 5 trials and finally set it to $0.4$.
For the negative constant reward in the decision-based attack model we tune it in the range of $-3$ to $-0.5$ for about 5 trials and set it to $-1$.
And the learning rate of $\mathbf{\vartheta}$ and $\mathbf{q}_w$ in the pre-trained attack model are tuned in the range of $1e-5$ to $1e-8$ and the range of $0.1$ to $0.5$ for about 10 trails, respectively. The learning rate of $\mathbf{\vartheta}$ is set to $1e-7$ and the learning rate of $\mathbf{q}_w$ is set to $0.3$.
Experimental results show that our attack model is not sensitive to hyper-parameters. In the hyper-parameter searching process, the attack success rates of our attack model fluctuate by about 5\%.
In addition, experimental results show that the variance between the success rates of our attack models on the valid set and on the test set is small (usually less than 2\%).

\begin{table}[!t]
\centering
\resizebox{\linewidth}{!}{
\begin{tabular}{c|c||c|c}
\toprule
\multicolumn{2}{c||}{Score-based Attacking} & \multicolumn{2}{c}{Decision-based Attacking}\\
\hline
Attack Model & \%Valid & Attack Model & \%Valid  \\ 
\hline
Original Sentence & 83   & Original Sentence & 83 \\ 
\hline
 GA+Embedding          & 57   &GA'+Embedding          & 60      \\
 RL+Embedding           & 57 &RL+Embedding           & 59        \\ 
  \hline
 PWWS+Synonym         & 51    & \multirow{2}{*}{AED} &  \multirow{2}{*}{47}     \\
 RL+Synonym         & 56     &      \\ 
 \hline
 PSO+Sememe          & 65   & PSO'+Sememe          & 58       \\  
 RL+Sememe          & 64    & RL+Sememe          & 62      \\  

\bottomrule
\end{tabular}
}
\caption{Human evaluation results of attack validity for score- and decision-based attack models on SST-2, where the second row additionally lists the evaluation results of original sentences.
``\%Valid'' refers to the percentage of instances that are annotated with the same sentiment as the ground truth of the corresponding original sentences.
}
\label{tab:valid}
\end{table}

\section{Human Evaluation of Attack Validity}
\label{sec:appendix}
We use human evaluation to evaluate the attack validity of different attack models.
Considering the cost, this evaluation is conducted on the SST-2 dataset only, where the victim model is ALBERT.
Specifically, we ask workers to annotate binary sentiments for $100$ randomly sampled adversarial examples and calculate the percentage of adversarial examples that are annotated with the same sentiment as the ground truth of original sentences. 
Each adversarial example is annotated by three workers and its final sentiment is decided by voting.
We also make annotations for the original sentences.

Table \ref{tab:valid} lists the validity results of different attack models in both score- and decision-based attack settings.
We also conduct significance tests to measure the difference between different models, and Table \ref{tab:valid_t_test_score} and \ref{tab:valid_t_test_decision} show the results.
With statistical significance threshold of \textit{p}-value of $0.05$, we can conclude that the attack validity of our model is comparable to the baseline methods (even better than AED in the decision-based attack setting).
In other words, although our model has higher attack success rates and attack efficiency, it does not sacrifice attack validity.
However, compared with original sentences, the adversarial examples generated by whichever attack model have obviously worse validity, which demonstrates that low attack validity is a common and unsolved problem for adversarial attacking.

\begin{table*}[!t]
\centering
\resizebox{.8\linewidth}{!}{
\begin{tabular}{cc||lcc}
\toprule
 Model 1 & Model 2 & \textit{p}-value & Significance    & Conclusion     \\ \hline
RL+Embedding                 & GA+Embedding        & 0.50                     & Not Significant & =              \\ \hline
                          RL+Synonym                  & PWWS+Synonym              & 0.24                    & Not Significant & =              \\ \hline
                          RL+Sememe                  &
                         PSO+Sememe     &   0.44 & Not Significant & = \\\hline
                          Original Sentence              & GA+Embedding        & 2.34e-5                 & Significant     & \textgreater{} \\ \hline
                          Original Sentence              &  RL+Embedding             & 2.34e-5                  & Significant     & \textgreater{} \\ \hline
                          Original Sentence              & PWWS+Synonym                  & 1.26e-5                  & Significant     & \textgreater{} \\ \hline
                          Original Sentence              & RL+Synonym                  & 4.11e-7                  & Significant     & \textgreater{} \\ \hline
                          Original Sentence              & PSO+Sememe                & 1.10e-3                  & Significant     & \textgreater{} \\ \hline
                          Original Sentence              & RL+Sememe                  & 1.78e-3                  & Significant     & \textgreater{} \\ \bottomrule

\end{tabular}
}
\caption{The Student's \textit{t}-test results of attack validity of different \textit{score}-based attack models, where ``='' means ``Model 1'' performs as well as ``Model 2'' and ``\textgreater{}'' means ``Model 1'' performs better than ``Model 2''.}
\label{tab:valid_t_test_score}

\end{table*}

\begin{table*}[!t]
\centering
\resizebox{.8\linewidth}{!}{
\begin{tabular}{cc||lcc}
\toprule
 Model 1 & Model 2 & \textit{p}-value & Significance    & Conclusion     \\ \hline
RL+Embedding & GA'+Embedding                       & 0.44                     & Not Significant & =              \\ \hline
                          RL+Sememe & PSO'+Sememe                  
                              &   0.28 & Not Significant & = \\\hline
                          RL+Embedding              & AED       & 0.04                & Significant     & \textgreater{} \\ \hline
                          RL+Sememe              &  AED            & 0.02                  & Significant     & \textgreater{} \\ \hline
                          Original Sentence              & GA'+Embedding        & 1.36e-4                 & Significant     & \textgreater{} \\ \hline
                          Original Sentence              &  RL+Embedding             & 7.71e-5                  & Significant     & \textgreater{} \\ \hline
                          Original Sentence              & PSO'+Sememe                & 4.29e-5                  & Significant     & \textgreater{} \\ \hline
                          Original Sentence              & RL+Sememe                  & 4.01e-4                  & Significant     & \textgreater{} \\ \hline
                          Original Sentence              & AED                  & 1.81e-8                  & Significant     & \textgreater{} \\ 
                          \bottomrule

\end{tabular}
}
\caption{The Student's \textit{t}-test results of attack validity of different \textit{decision}-based attack models, where ``='' means ``Model 1'' performs as well as ``Model 2'' and ``\textgreater{}'' means ``Model 1'' performs better than ``Model 2''.}
\label{tab:valid_t_test_decision}

\end{table*}
\section{Runtime}
We use a single 2080-Ti GPU for each attacking experiment. The time costs of querying ALBERT on SST-2, XLNet on AGNews and RoBERTa on MNLI are 0.05, 0.03 and 0.03 second for one query, respectively. The time costs of the attacking experiments mainly consist of querying victim models and can be estimated by the number of queries.